\documentclass[10pt,twocolumn,letterpaper]{article}

\usepackage{iccv}
\usepackage{times}
\usepackage{epsfig}
\usepackage{graphicx}
\usepackage{amsmath}
\usepackage{amssymb}

\usepackage{booktabs}
\usepackage{subcaption}
\usepackage[accsupp]{axessibility} 

\usepackage[breaklinks=true,bookmarks=false]{hyperref}
\usepackage{cleveref}

 \iccvfinalcopy 

\def\iccvPaperID{9754} 

\ificcvfinal\fi

\begin{document}

\title{Probabilistic Triangulation for Uncalibrated Multi-View 3D Human Pose Estimation}

\author{Boyuan Jiang\textsuperscript{1,2}, Lei Hu\textsuperscript{1,2}\,  Shihong Xia\textsuperscript{1,2}\thanks{Corresponding author}\\ 
$^1$Institute of Computing Technology, Chinese Academy of Sciences;\\ $^2$University of Chinese Academy of Sciences\\
{\tt\small \{jiangboyuan20s, hulei19z, xsh\}@ict.ac.cn}
}

\maketitle
\ificcvfinal\thispagestyle{empty}\fi

\begin{abstract}
3D human pose estimation has been a long-standing challenge in computer vision and graphics, where multi-view methods have significantly progressed but are limited by the tedious calibration processes. Existing multi-view methods are restricted to fixed camera pose and therefore lack generalization ability. This paper presents a novel Probabilistic Triangulation module that can be embedded in a calibrated 3D human pose estimation method, generalizing it to uncalibration scenes. The key idea is to use a probability distribution to model the camera pose and iteratively update the distribution from 2D features instead of using camera pose. Specifically, We maintain a camera pose distribution and then iteratively update this distribution by computing the posterior probability of the camera pose through Monte Carlo sampling. This way, the gradients can be directly back-propagated from the 3D pose estimation to the 2D heatmap, enabling end-to-end training. Extensive experiments on Human3.6M and CMU Panoptic demonstrate that our method outperforms other uncalibration methods and achieves comparable results with state-of-the-art calibration methods. Thus, our method achieves a trade-off between estimation accuracy and generalizability. Our code is in https://github.com/bymaths/probabilistic\_triangulation
\end{abstract}

\section{Introduction}

3D human pose estimation is a fundamental tool in many downstream 
 applications of computer vision and computer graphics, such as motion recognition\cite{du2017rpan}, human-computer interaction\cite{munea2020progress}, movie game animation\cite{borodulina2019application}, virtual reality\cite{obdrvzalek2012real}, etc.
The development of these fields requires a more accurate, convenient, and robust 3D human pose estimation algorithm.



Recently, RGB-based pose estimation has become the trend because of its convenience, wide application scenarios, and low cost\cite{regazzoni2014rgb}. These methods can be divided into two categories: single-view and multi-view methods. Although single-view methods\cite{zhu2022motionbert,zhang2022mixste,cheng20203d,zheng20213d,pavllo20193d} have more widespread application scenarios, the lack of depth information makes single-view estimation of 3D human pose an ill-posed problem, especially when dealing with occlusions and multi-person scenes\cite{liu2022recent}. Multi-view methods\cite{iskakov2019learnable,he2020epipolar,zhang2021direct,tu2020voxelpose,bartol2022generalizable,gordon2021flex,takahashi2018human,lee2022extrinsic} can achieve more accurate results than single view-based methods because multi-view cameras can restore the depth information as well as the occluded parts.

However, the multi-view approach faces several issues when applied in practice. First, multi-view methods require synchronized and calibrated cameras, which complicates the process. Second, existing datasets for multi-view human pose estimation lack the diversity of camera external parameters. This results in the trained model cannot generalize well to unseen camera parameters, thus requiring re-annotation and re-training once the camera configurations change.

To alleviate the above issues, we propose Probabilistic Triangulation module for uncalibrated 3D human pose estimation. Our module enables the network to get rid of camera pose parameters during the training process, which results in better generalization performance to be applied to wild scenes. The main idea is to use a probability distribution to model the camera pose, making the whole process differentiable. Specifically, we transform the uncalibrated 3D human pose estimation into a nonlinear optimization problem based on the reprojection error. The camera pose distribution is estimated from the 2D pose features, and then the posterior probability of the camera pose is computed by Monte Carlo sampling to iteratively update this distribution.

We conducted extensive experiments on two benchmark datasets (Human3.6M\cite{ionescu2013human3} and CMU Panoptic\cite{joo2015panoptic,simon2017hand,xiang2019monocular}) to demonstrate that our method outperforms other uncalibrated methods and achieves comparable results with the state-of-the-art calibration methods, which indicates that our method improves the generalized ability to camera parameters in multi-view human pose estimation while maintaining high accuracy. 

In summary, our contributions are mainly three-fold.
\begin{itemize}
    \item We propose Probabilistic Triangulation that can generalize the calibrated 3D human pose estimation algorithm to uncalibrated scenes to estimate 3D human pose in the absence of camera pose.
    \item We propose an approximation method for camera pose distribution and an updated distribution method based on Monte Carlo sampling. It is used to model the camera pose of uncalibrated scenes with accuracy and robustness beyond classical algorithms.
    \item We conducted extensive experiments on two benchmark datasets (Human3.6M and CMU Panoptic) to demonstrate the accuracy, generalization capability, and applicability of Probabilistic Triangulation to wild scenes.
\end{itemize}
\section{Related work}
\subsection{Calibrated 3D pose estimation}
3D human pose estimation is a vital problem in computer vision and graphics, where the single view-based method has been a research spotlight due to the convenience and comprehensive application scenarios of a monocular camera. To alleviate the ambiguity problem caused by occlusion and the lack of depth information, many single view-based  works\cite{zhang2022mixste,cheng20203d,zheng20213d,pavllo20193d,belagiannis20143d} have tried to introduce the temporal information or body-related prior during training.

Multi-view images retain more information compared to single-view ones. The association of information between different views eliminates most of the ambiguity. So it is easier to achieve high-accuracy 3D human pose estimation in multi-view settings. Fusing the information of different views is the core problem of multi-view human pose estimation. Existing multi-view methods include two processes, 2D pose estimation and 3D pose reconstruction. The former consists of a pre-trained 2D pose estimator\cite{chen2018cascaded,cao2017realtime}, and multi-view information fusion occurs in the 3D pose reconstruction. Multi-view methods are divided into two categories according to the different ways of representing human pose. The first is the point-based method\cite{qiu2019cross,bultmann2021real,dong2019fast}, which uses the spatial coordinates of the joint points as the human pose representation. This representation makes it easier to introduce temporal and body information explicitly. The second is the heatmap-based method, which uses the heatmap as the human pose representation\cite{he2020epipolar,zhang2021direct,iskakov2019learnable}, and learnable triangulation represents this method\cite{iskakov2019learnable,tu2020voxelpose}. The point representation is more concise and explicit but loses information, while the heatmap representation retains more information but is not easily bounded explicitly. Our proposed Probabilistic Triangulation is applicable in both representations.

\begin{figure*}
  \centering
    \includegraphics[width=17cm]{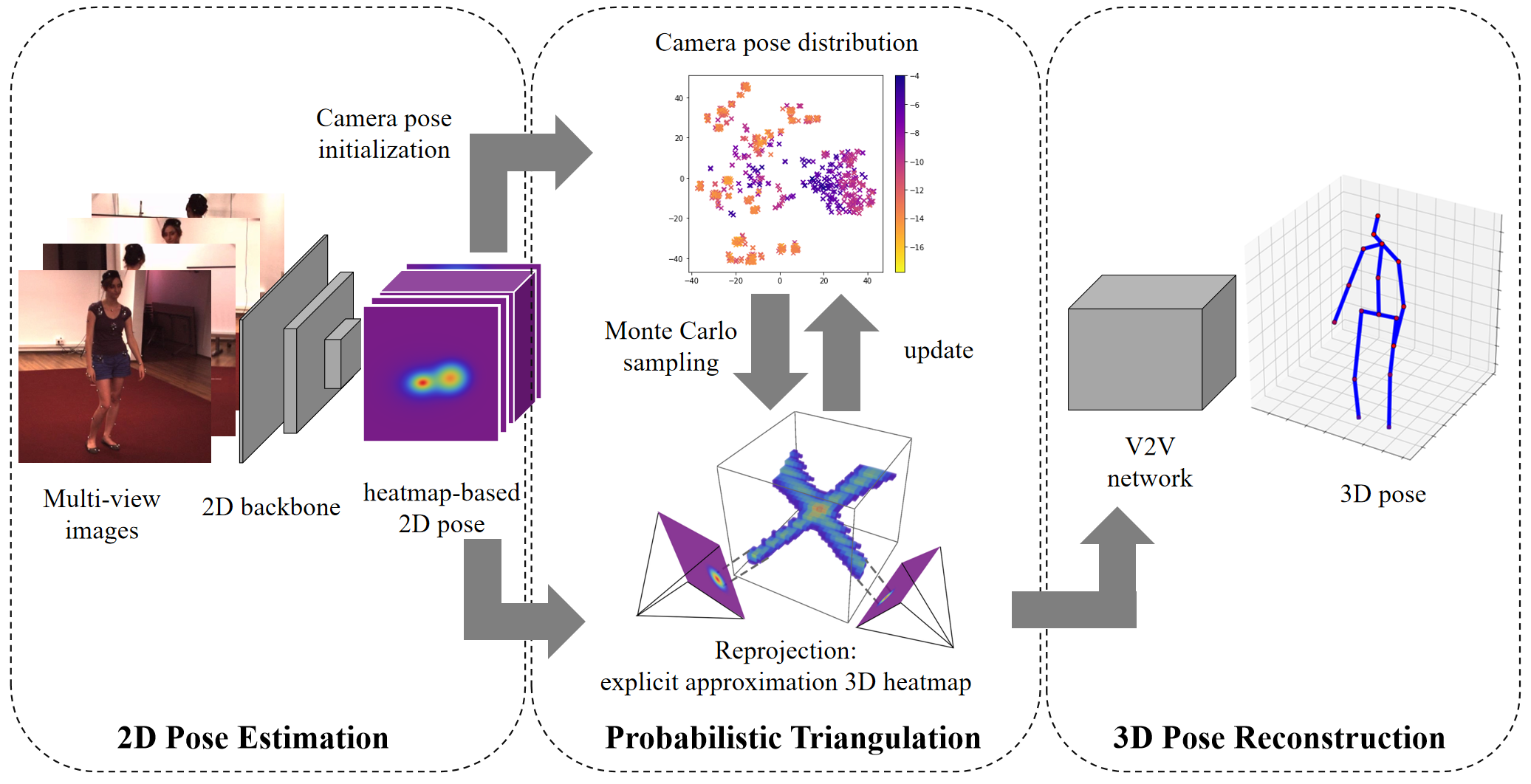}
  \caption{Probabilistic Triangulation for uncalibrated 3D human pose estimation pipeline. The inputs to the network are images from different views at the same moment. The 2D human pose heatmaps are estimated from the images and used to initialize the camera pose distribution. Then, Monte Carlo sampling is performed iteratively and each sampled camera parameter is computed by reprojection to obtain an explicit approximation of the 3D heatmap. The 3D heatmap calculates the weights of the sampled camera parameters. These weights are then used to update the pose distribution of the cameras. The weighted 3D heat map is used as input for the subsequent 3D pose reconstruction, for estimating the 3D human pose.}
  \label{fig:pipeline}
\end{figure*}

\subsection{Uncalibrated 3D pose estimation}
Uncalibrated 3D pose estimation is a relatively new field\cite{takahashi2018human,bartol2022generalizable,gordon2021flex,lee2022extrinsic}. Takahashi's work\cite{takahashi2018human} is a point-based method that establishes explicit constraints based on reprojection errors and the human body prior to transform the problem into an optimization problem and iteratively solve for the optimal solution. Lee's work\cite{lee2022extrinsic} is also a point-based method, estimating the camera pose using the 3D pose estimated from different views. FLEX\cite{gordon2021flex} estimates a time-invariant bone length while estimating joint rotations frame by frame and reconstructing 3D pose by forward kinematics. In this approach, the bone length is a link to fuse the multi-view information, and the rotation estimated for each view implicitly constrains the camera pose. However, this method cannot accurately estimate the camera pose due to the lack of transformation of the global coordinate system. Therefore, the output 3D pose is camera dependent. Bartol et al. \cite{bartol2022generalizable} use a scoring network to evaluate the weight of the current proposal's camera pose or 3D human pose, and the final weighting is used to obtain a definite result. This method does not directly estimate the 3D human pose for uncalibrated scenes, but it can estimate the camera pose before and use other calibration methods to estimate the 3D human pose. This method does not directly estimate the 3D human pose for uncalibrated scenes, but it can estimate the camera pose before using other calibration methods to estimate the 3D human pose. Because the experiments for 3D human pose estimation in this work are for calibrated scenes, they are placed in the calibration method in the comparison experiments later in the paper.

Current methods are limited to point-based representation. The heatmap-based representation retains more information than the point-based representation. This is because the accuracy of existing 2D human pose estimators is not sufficient for direct calibration. Small errors in points can be rapidly amplified by the camera pose estimation process, making the method ineffective. Our proposed camera pose distribution can effectively increase the robustness of the camera estimation process and support heatmap-based representation.

\subsection{Probabilistic camera pose estimation}

Camera pose estimation is a fundamental problem in computer vision\cite{hartley2003multiple}. PNP methods solve the case of known 3D-2D point pairs, while triangulation solves the case of known 2D-2D point pairs. Bundle adjustment transforms the camera matching problem into an optimization problem starting from reprojection errors\cite{agarwal2015others}. However, matching feature points and solving them using classical methods does not apply to all scenarios. In the field of object and camera pose estimation, many works try to improve the network architecture that allows an end-to-end approach for pose estimation\cite{chen2021monorun,chen2020end}. The work EPro-PnP by Chen et al\cite{chen2022epro}. proposes a probabilistic PnP layer that allows the gradient back-propagated through the pose. 

Inspired by EPro-PnP, we use probability distributions to model camera pose. Compared with the camera pose represented by a single variable, the camera pose represented by the distribution has more robust performance and effectively mitigates the effect of 2D human pose detection errors.
\section{Method}
In this paragraph, we will review the multi-view uncalibrated 3D human pose estimation problem and introduce our Probabilistic Triangulation module. The overall pipline is shown in Figure \ref{fig:pipeline}.

\begin{figure*}
  \centering
  \begin{subfigure}{0.33\linewidth}
    \includegraphics[width=5cm]{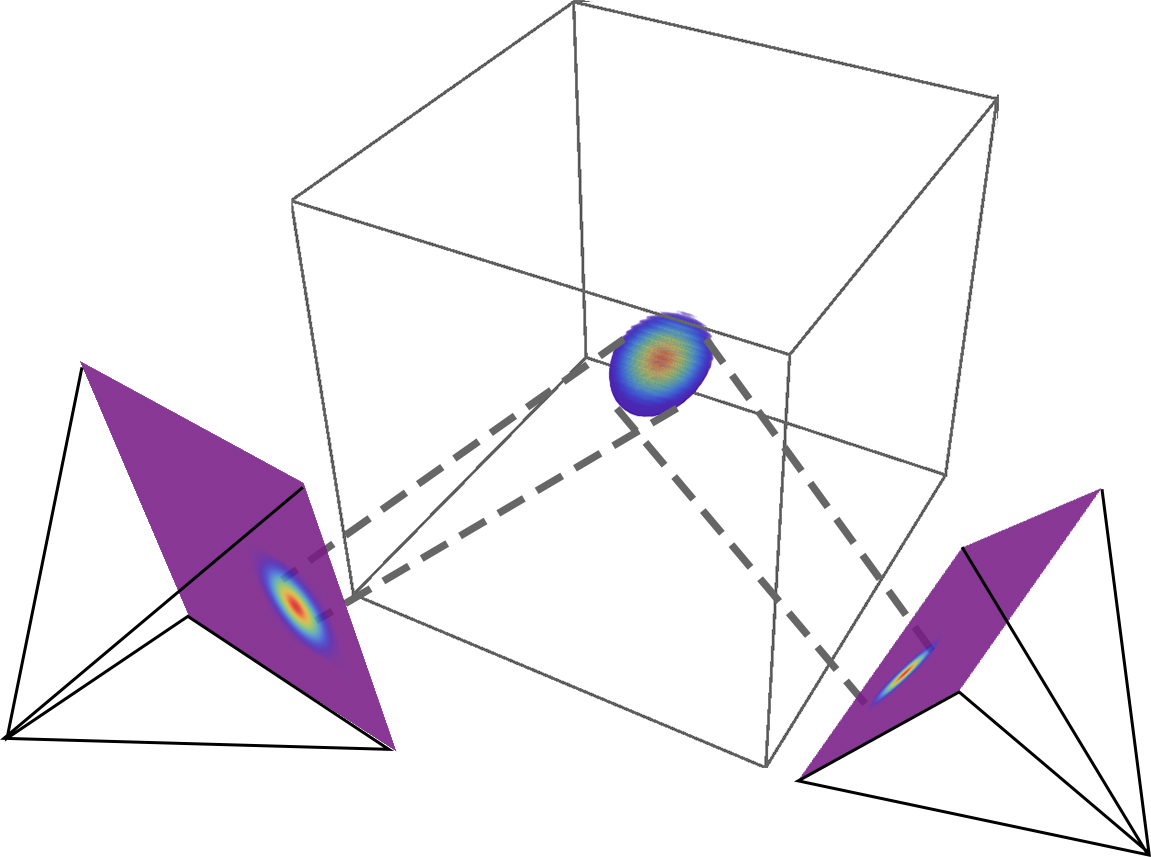}
    \caption{Implicit continuous.}
    \label{fig:short-a}
  \end{subfigure}
  \hfill
  \begin{subfigure}{0.33\linewidth}
    \includegraphics[width=5cm]{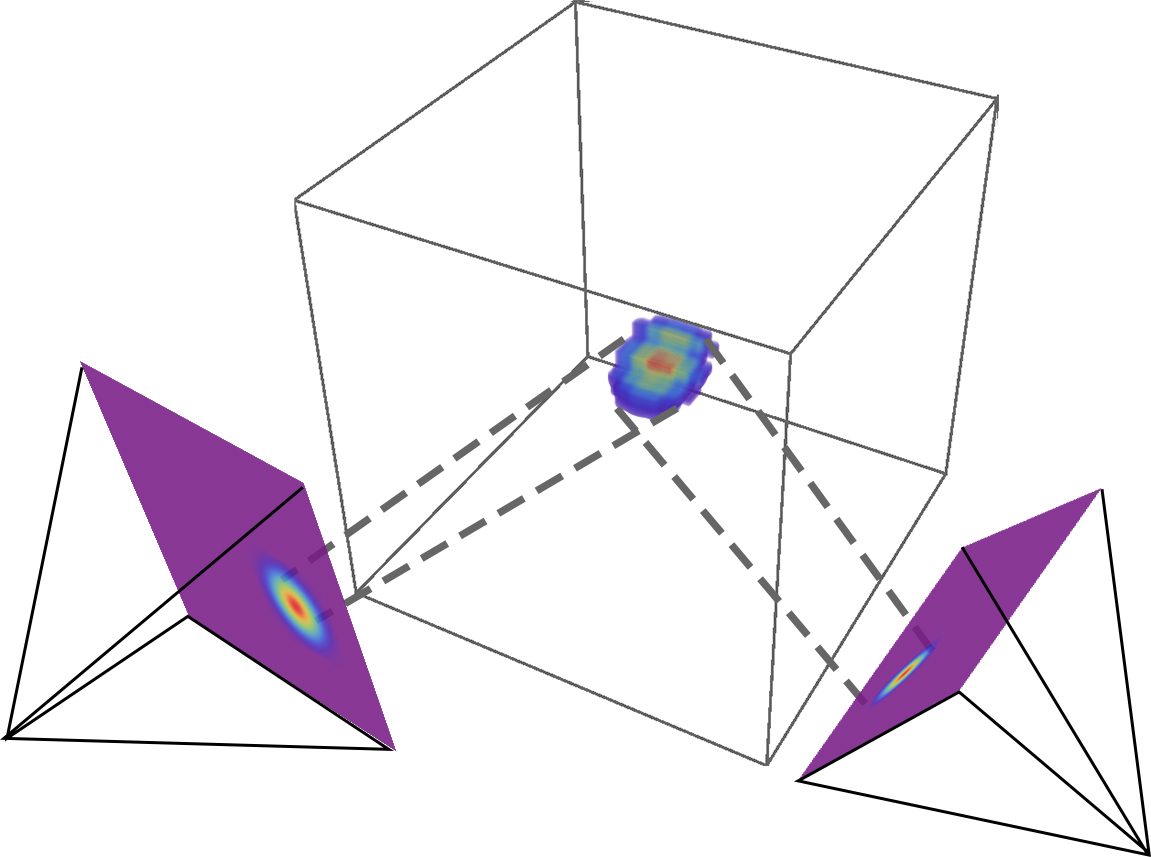}
    \caption{Implicitly discrete.}
    \label{fig:short-b}
  \end{subfigure}
  \hfill
  \begin{subfigure}{0.33\linewidth}
    \includegraphics[width=5cm]{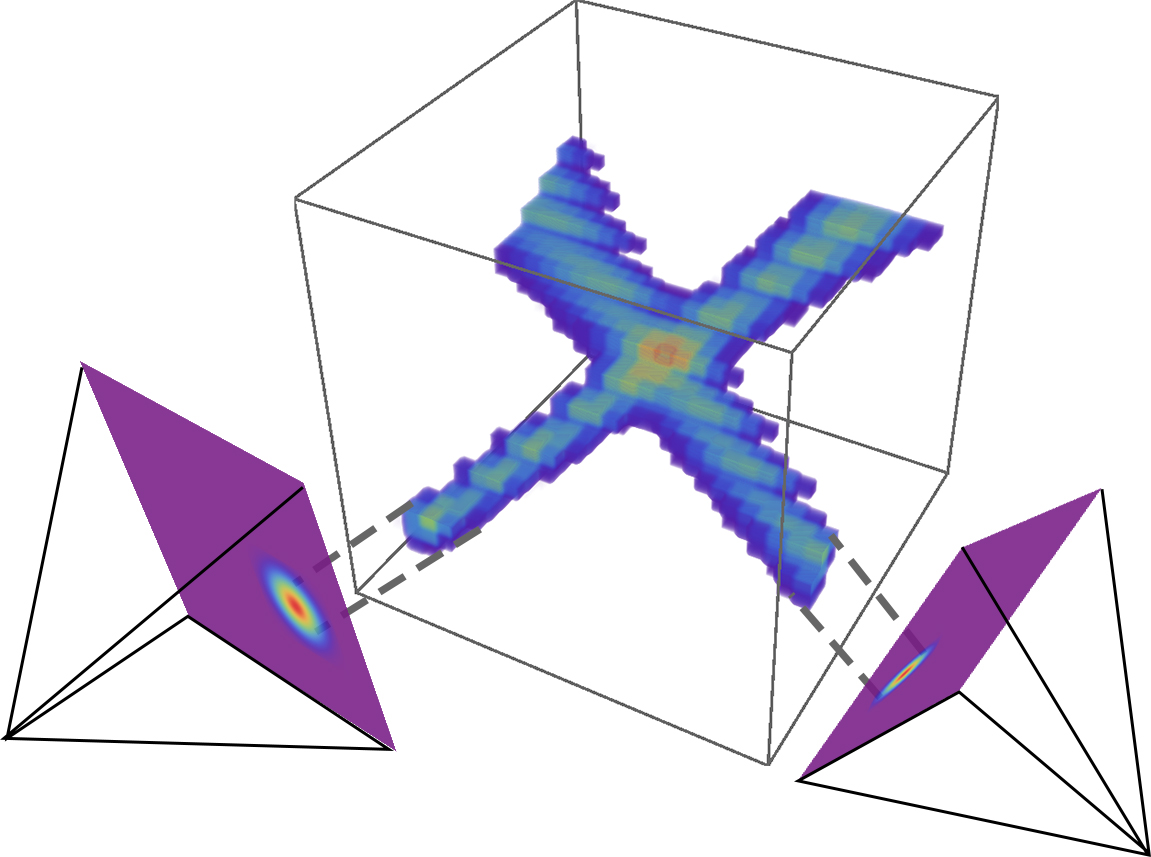}
    \caption{Explicit discrete approximation.}
    \label{fig:short-c}
  \end{subfigure}
  \caption{3D heatmap under different definitions}
  \label{fig:short}
\end{figure*}

\subsection{Uncalibrated 3D human pose estimation}

Given that the images $I \in \mathbb{R}^{K\times H \times W}$ from $K$ views at a specific moment, our goal is to estimate the 3D pose $X^{3D} $ of the target human from it. Given the camera pose $y = \{y_k\in SE(3)| k=1 \dots K\}$, the optimal $X^{3D}$ will make the reprojection error (3D$\rightarrow 2D$) small enough and fits the prior distribution of the human pose learned in the dataset. Therefore, the estimation algorithm is often divided into two parts, 2D pose estimation, and 3D pose reconstruction. In 2D pose estimation, a neural network $M_1$ is employed to estimate the 2D pose $X^{2D} = M_1(I)$ from each view. In the 3D pose reconstruction, the $X^{2D}$ of different views are fused using a triangulation or a learnable triangulation method $M_2$ to obtain $X^{3D} = M_2(X^{2D},y)$.

Based on the reprojection error, the calibrated 3D human pose estimation is to solve an optimization problem:

\begin{equation}
\begin{aligned}
&\min_{X^{3D}} \quad E\left[proj(X^{3D},y)- X^{2D}\right]
\end{aligned}
\label{eq:opt}
\end{equation}

where $proj(\cdot)$ denotes the projection function. The neural network-based algorithm replaces $X^{3D}$ in it with $M_2(X^{2D},y)$, transforming the original optimization problem into solving for the optimal $M_2(\cdot)$ network parameters.

When the camera pose $y$ is an unknown parameter, $y$ also becomes a variable to be optimized. Uncalibrated 3D human pose estimation can be similarly transformed into an optimization problem:

\begin{equation}
\begin{aligned}
&\min_{X_{3D},y} \quad E\left[proj(X^{3D},y)- X^{2D}\right]\\
\end{aligned}
\label{eq:opt2}
\end{equation}

This problem is tough to solve and exists multiple solutions. There are three main reasons for this:
\begin{enumerate}
    \item If an iterative method is used to optimize $y$ and $X^{3D}$ alternately, the results tend to fall into local optimum.
    \item If a neural network is used to regress $y$ and then solve for $X^{3D}$, it will be constrained by the lack of diversity in the existing data set $y$ to train.
    \item If the neural network is used to regress $X^{3D}$ directly from the monocular view and then to solve $y$, it will introduce the ill-conditioned problem of monocular scale ambiguity.
\end{enumerate}

In summary, we have to extract more information from the only known quantity $X_{2D}$ to obtain a relatively accurate estimate of $y$, i.e., a valid estimate $p(y|X^{2D})$.

We use camera pose distribution approximation and Monte Carlo sampling to solve this problem. (see Probabilistic Triangulation part of Figure \ref{fig:pipeline}) Specifically, we maintain a proposal distribution $q(y)$ for approximating the posterior distribution $p(y|X^{2D})$ of camera poses during one inference. Monte Carlo sampling is continuously performed on $q(y)$, and the weights are calculated from the sampled points, and then $q(y)$ is updated according to the weights. After several iterations, the results generated from the sampled points are weighted and averaged as the input to the 3D pose reconstruction network. We refer to this process as Probabilistic Triangulation. Figure \ref{fig:pipeline} shows the pipeline of uncalibrated 3D human pose estimation after adding Probabilistic Triangulation.

The Probabilistic Triangulation can be flexibly embedded into any kind of calibrated 3D human pose estimation, reducing its dependence on camera pose for realistic scenarios. The differentiable computational process makes it possible to connect the 2D pose estimation and 3D pose reconstruction parts, allowing end-to-end training.


\subsection{Probabilistic Triangulation}\label{sec3.2}

We detail the Probabilistic Triangulation using the pipeline in Figure \ref{fig:pipeline} as an example. In subsequent representations, $y$ is used to denote the camera pose and $y_{gt}$ the ground truth of the camera pose. The 2D human pose is represented as $X^{2D} = \{ x_{i}^{2D} \in \mathbb{R}^{K \times H'\times W'}|i=1\dots N \}$, and the 3D human pose is represented as $ X^{3D} = \{x_{i}^{3D} \in \mathbb{R}^{K\times L_1\times L_2 \times L_3}|i=1\dots N\}$, where $H'$ and $W'$ denote the height and width of the 2D heatmap, and $L_1,L_2,L_3$ denote the length, width, and height of the 3D heatmap.

\textbf{Camera pose distribution approximation} In order to solve the optimization problem \ref{eq:opt2} with multiple solutions., we first normalized the multi-view camera pose to limit the range of solutions. Specifically, through an affine transformation of the whole coordinate system, the rotation matrix of the first view is changed into a unit matrix, the translation vector becomes a zero vector, and the translation vector of the second view has a modulus of one. The above affine changes can guarantee the uniqueness of the solution.

Each camera pose is represented as a $7$-dimensional vector, where 3D rotations are represented using unit quaternions and 3D translations are represented using 3D vectors. To better express the camera distribution, the angular central Gaussian distribution is introduced to represent the 3D rotation \cite{tyler1987statistical}, and the multivariate t-distribution is introduced to represent the 3D translation. The $p(y|X^{2D})$ is calculated for each sampling point $y$ to obtain the covariance matrix of the angular central Gaussian distribution, and the mean and covariance matrix of the multivariate t-distribution, which in turn updates the parameters. The initial parameters of the distributions are first estimated based on the 2D human pose using the eight-point method. Then, a portion is sampled around the initial value and another portion is sampled around the entire camera domain. The distribution is initialized by the sampling result. Since the input of a single frame may cause a large error in the eight-point method, and initializing the distribution in this way increases the robustness of the algorithm.

\textbf{Monte Carlo sampling} We assume that the 2D heatmap is the marginal distribution of the 3D heatmap along the projection direction, and the problem can be transformed into an optimization problem based on the reprojection error:

\begin{equation}
\begin{aligned}
&\min_{y,X^{3D}} \quad \sum_i^{N}
\int_{\sigma_i} \left\| x^{2D}_i(\sigma_i) - \int_{proj(z,y_i)=\sigma_i}x^{3D}(z)\, \mathrm{d} z \right\|_2 \,\mathrm{d}\sigma_i
\end{aligned}
\end{equation}
where $z$ denotes the body element in 3D space, $\sigma_i$ denotes the area element of the $i$th camera plane, and $proj(\cdot)$ denotes the projection function. A more general heatmap representation is adopted here, where $x^{3D}$ and $x^{2D}$ are considered as the probability density functions of the heatmap distribution. In practice, the heatmap is often stored in a discrete way, and the integrals need to be discretized as well. However, it is difficult to establish the constraints between $y$ and $X^{3D}$ explicitly, making the above optimization problem difficult to solve.


Therefore, we use an approximate 3D heatmap explicitly defined as:

\begin{equation}
\begin{aligned}
 x^{3D}(z) = \frac{1}{N}\sum_{\substack{i \\ proj(z,y_i) =
\sigma_i }}^N x_i^{2D} (\sigma_i)\\
x^{2D}_i(\sigma_i) = \max_{proj(z,y_i)=\sigma_i}{x^{3D}(z)}
\end{aligned}
\end{equation}

This definition is similar to the volume of the learnable triangulation, with the advantage that $X^{3D}$ can be computed explicitly and differentiably from $y$ and $X^{2D}$. It also has some physical meaning as far as the location of the extreme points taken is similar to the original representation. The visualization of the 3D heatmap under different definitions is shown in Figure \ref{fig:short}.

The generated $X^{3D}$ retains the information of the heatmap and can be used as input for subsequent networks. Thus, the problem is simplified to:

\begin{equation}
    \min_y \sum_i^N 
 \|
 \underbrace{
 \sum_z 
 \left(x^{3D}(z) x_i^{2D}(\sigma_i)\right)^2 \left( x_i^{2D}(\sigma_i)  - 
 x^{3D}(z) \right)
 }_{f_i(y)}
 \|_2
\end{equation}

The reprojection error is employed as the negative logarithm of the likelihood function $p(X_{2D} | y)$.

\begin{equation}
    p(X^{2D}|y) =\exp -\sum_i^N \|f_i(y) \|_2
\end{equation}

Suppose that the camera pose $p(y)$ is subject to a uniform distribution, $p(y|X^{2D})$ can be calculated by Bayesian formula:

\begin{equation}
    p(y|X^{2D}) = \frac{p(X^{2D}|y)}{E\left[p(X^{2D}|y)\right]}
\end{equation}

The calculation of denominator expectation can be estimated by Monte Carlo sampling approximation. The specific calculation formula is as follows.

\begin{equation}
    E\left[ p(X^{2D}|y)\right] \approx  \frac{1}{M} \sum_j^{M} q(y_j)p(X^{2D}|y),y_j \sim q(y)
\label{eq:monte carlo}
\end{equation}
where $M$ denotes the total number of samples. Then, $q(y)$ is updated using the computed $p(y|X^{2D})$.

Finally, the normalized $q(y)$ is used as the weight and the $X^{3D}$ generated by all sampled points $y$ is weighted and averaged as the input to the subsequent 3D pose reconstruction network.

\begin{equation}
    X^{3D} = \frac{\sum_j q(y_j) X^{3D}_j }{\sum_j q(y_j)}
\end{equation}

\begin{table*}
\small
  \centering
  \tabcolsep=0.08cm
  \begin{tabular}{c|ccccccccccccccc|c}
    \toprule

    Method &
    Dir. & Disc.& Eat & Greet & Phone & Photo & Pose & Purch. & Sit & SitD. & Smoke & Wait &  WalkD. & Walk  & WalkT. & Mean \\
    \midrule
    \midrule
    Calibration method\\
    \midrule
    Remelli et al.\cite{remelli2020lightweight} & 27.3 &32.1 &25.0 &26.5 &29.3 &35.4 &28.8 &31.6 &36.4 &31.7 &31.2& 29.9 &26.9 &33.7 &30.4 &30.2\\
    Bartol et al.\cite{bartol2022generalizable}
    & 27.5 & 28.4 & 29.3 & 27.5 & 30.1 & 28.1 & 27.9 & 30.8 & 32.9 & 32.5 & 30.8 & 29.4 & 28.5 &  30.5 & 30.1 & 29.1\\
    He et al. \cite{he2020epipolar} &25.7 &27.7 &23.7 &24.8& 26.9 &31.4 &24.9 &26.5 &28.8 &31.7 &28.2 &26.4& 23.6 &28.3 &23.5 &26.9\\
    Qiu et al. \cite{qiu2019cross} & 24.0 &26.7 &23.2 &24.3 &24.8 &22.8& 24.1& 28.6 &32.1 &26.9 &31.0 &25.6 &25.0 &28.0 &24.4 &26.2\\
    Ma et al. \cite{ma2021transfusion} & 24.4& 26.4 &23.4 &21.1 &25.2 &23.2 &24.7& 33.8& 29.8 &26.4 &26.8& 24.2& 23.2 &26.1 &23.3 &25.8\\
    Iskakov et al.\cite{iskakov2019learnable} &
    19.9 & 20.0 & 18.9 & 18.5 & 20.5 & 19.4 & 18.4 & 22.1 & 22.5 & 28.7 & 21.2 & 20.8 & 19.7 & 22.1 & 20.2 & 20.8\\
    \midrule
    \midrule
    Uncalibration method\\
    \midrule
    Gordon et al.\cite{gordon2021flex} 
    & \textbf{22.0} & \textbf{23.6} & \textbf{24.9} & \textbf{26.7}  & \textbf{30.6}
    & 35.7 & 25.1 & \textbf{32.9} & \textbf{29.5} & 32.5 & 32.6 & 26.5 & 34.7 & 26.0 & 27.7 & 30.2\\
    
    Ours method&24.0&25.4&26.6&30.4&32.1&\textbf{20.1}&\textbf{20.5}&36.5&40.1&\textbf{29.5}&\textbf{27.4}&\textbf{27.6}&\textbf{20.8}&\textbf{24.1}&\textbf{22.0}&\textbf{27.8}\\
    \bottomrule
  \end{tabular}
  \caption{Results of the evaluation on the Human3.6M dataset. The table shows the MPJPE in millimeters for the published state-of-the-art calibrated and uncalibrated methods.}
  \label{tab:1}
\end{table*}

\subsection{Training Loss}
The entire training process include two loss terms: $\mathcal{L}_{3d}$ and $\mathcal{L}_{cam}$.

We measure the difference between the true distribution of camera pose $z(y)$ and the posterior distribution $p(y|X^{2D})$ using KL divergence.

\begin{equation}
\begin{aligned}
\mathcal{L}_{cam} & = D_{KL}(z(y) \| p(y|X^{2D}))\\
=& - \frac{1}{2} \sum_i^{N} \| f_i(y_{gt})\|_2 + \log E\left[p(X^{2D}|y) \right]
\end{aligned}
\end{equation}
where the expectation of the second half is also obtained by the calculation of the equation \ref{eq:monte carlo}. 

The 3D human pose estimation loss is the same as previous works\cite{iskakov2019learnable}, which encourages the predicted 3d poses to be close to the ground truth. The formula can be summarized as follow:
\begin{equation}
 \mathcal{L}_{3d} =  \| \mathrm{softargmax}(X^{3D}) - X_{gt} \|_1 - \beta \cdot \log(X^{3D}(X_{gt}))
\end{equation}

In end-to-end training, we simultaneously optimize these two loss terms to train the networks, and the total loss function can be written as:
\begin{equation}
    \mathcal{L}_{total} = \lambda_1 \mathcal{L}_{cam} + \lambda_2 \mathcal{L}_{3d}
\end{equation}
where the $\lambda_1 = 1 $ and $\lambda_2 = 0.1 $ in our experiment setting. 
\section{Experiments}

\subsection{Dataset \& Metrices}

We conducted experiments on two of the most commonly used 3D human pose datasets available today, Human3.6M\cite{ionescu2013human3} and CMU Panoptic\cite{joo2015panoptic,simon2017hand,xiang2019monocular}. We used the 17-joint MPJPE (Mean Per Joint Position Error) as a metric to evaluate our method. In training, we normalize the ground truth according to the method in section\ref{sec3.2} to ensure the uniqueness in the camera pose solution. In validation, the estimated 3D human pose is denormalized to the original coordinate system to ensure the correctness of the MPJPE calculation.

\subsection{Implementation details}

We added the Probabilistic Triangulation based on the work learnable triangulation by Iskakov et al.\cite{iskakov2019learnable} to realize uncalibrated 3D human pose estimation. The same experimental parameter settings as the learnable triangulation method were used. As with learnable triangulation, our method uses predictions from the algebraic triangulation. Our method can also be fine-tuned and be trained in an end-to-end manner. $q(y)$ is initialized using a uniform distribution. The number of iterations is four, and the number of Monte Carlo samples is 256.

\subsection{Comparisons with State-of-the-art}

\begin{table}
\small
  \centering
  \tabcolsep=0.1cm
  \begin{tabular}{cc}
    \toprule
    Method & MPJPE,mm\\
    \midrule
    \midrule
    Calibrated method\\
    \midrule
    Iskakov et al. Algebraic\cite{iskakov2019learnable} & 21.3\\
    Iskakov et al. Volumetric\cite{iskakov2019learnable} & 13.7\\
    \midrule
    \midrule
    Uncalibrated method\\
    \midrule
    Bartol et al. \cite{bartol2022generalizable} & 25.4\\
    Our method & 24.2\\
    \bottomrule
  \end{tabular}
  \caption{Results of the MPJPE on the CMU Panoptic dataset ( using four cameras).}
  \label{tab:2}
\end{table}

We compared the existing state-of-the-art calibrated and uncalibrated 3D pose estimation algorithms. The evaluation results on the Human3.6M dataset and the CMU Panoptic dataset are shown in Table \ref{tab:1} and Table \ref{tab:2}, respectively. The experimental results show that the MPJPE of our method has outperformed the state-of-the-art uncalibrated approaches and achieved comparable performance to calibrated methods. It is worth noting that our method achieves better results in most scenarios of uncalibrated setting (see Table \ref{tab:1}), but the results are poor in the Purch. and Sit. sub-datasets. This is because there are more occlusions in these two subsets, resulting in the 2D joint estimation being inaccurate, which further affects the estimation of the camera parameter distribution. Therefore, the final accumulated error affects the 3D human pose estimation performance. Inferring the camera parameters from the features of the detected human inevitably leads to estimation results depending on the capability of the human feature detector. However, our method still outperforms existing uncalibrated methods in terms of overall MPJPE for both datasets.

\begin{figure*}
  \centering
  \begin{subfigure}{0.33\linewidth}
  \centering
    \includegraphics[width=4cm]{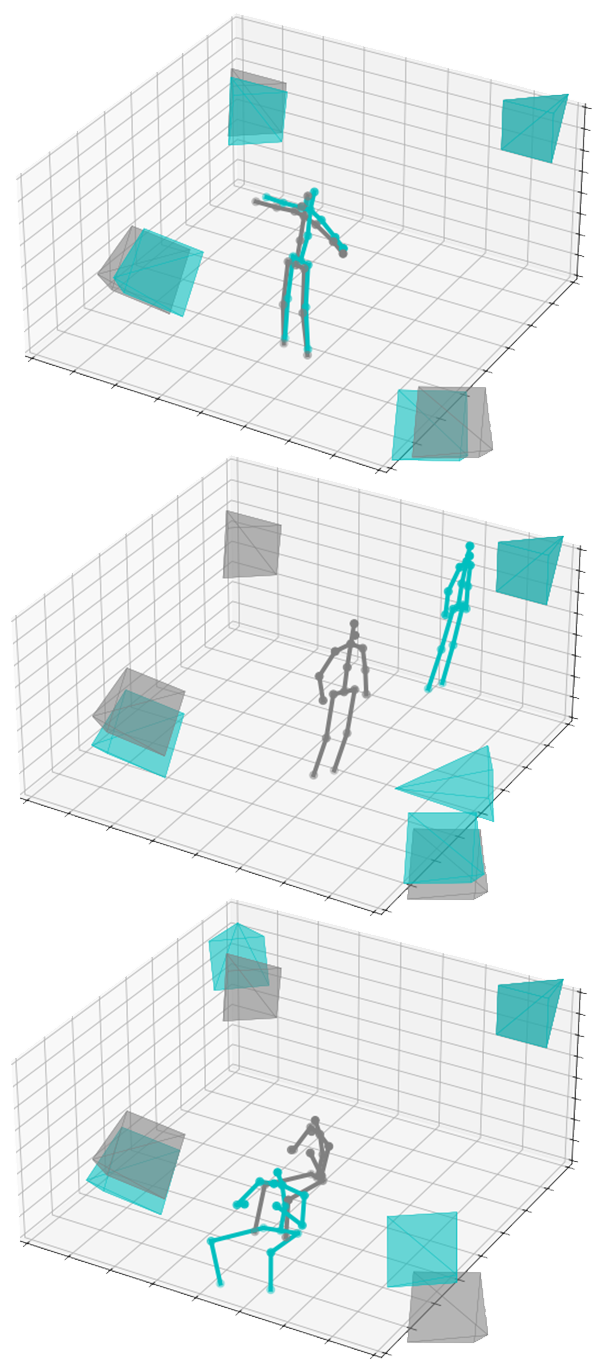}
    \caption{Bundle Adjustment.}
    \label{fig:short-a1}
  \end{subfigure}
  \hfill
  \begin{subfigure}{0.33\linewidth}
  \centering
    \includegraphics[width=4cm]{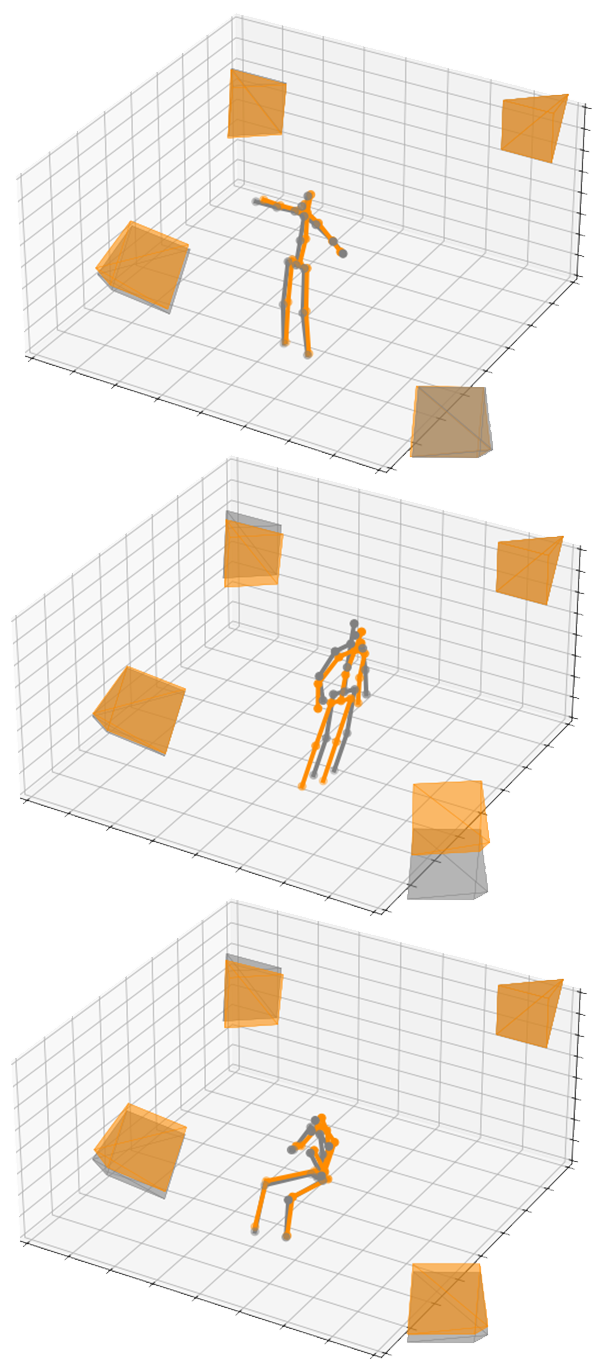}
    \caption{RANSAC 8-point algorithm.}
    \label{fig:short-b1}
  \end{subfigure}
  \hfill
  \begin{subfigure}{0.33\linewidth}
  \centering
    \includegraphics[width=4cm]{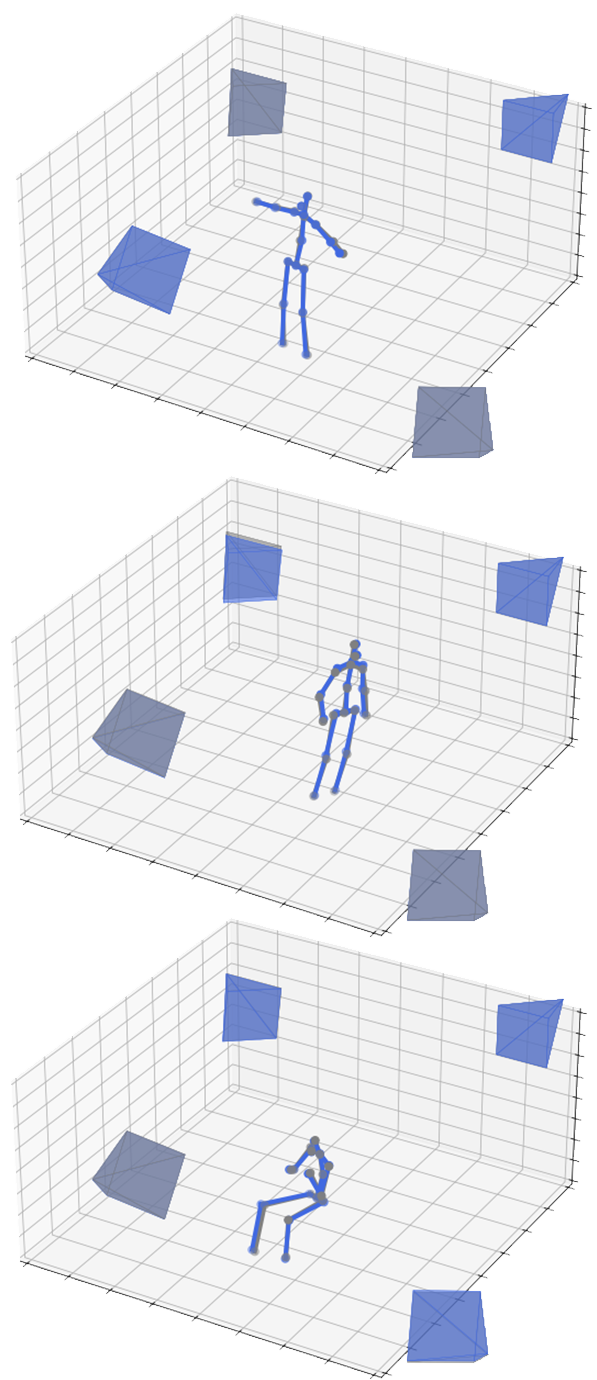}
    \caption{Ours method.}
    \label{fig:short-c1}
  \end{subfigure}
  \caption{Visualization of the camera pose parameters and 3D human pose obtained by different algorithms. The quadrilateral cone indicates the camera pose parameters, the gray part indicates the ground truth, and the colored part indicates the prediction results of different algorithms.}
  \label{fig:3}
\end{figure*}

\subsection{Evaluation on Camera Pose Estimation}

To evaluate the capability of our camera pose estimation, we have conducted comparisons with some classical camera parameter estimation algorithms on the Human3.6M dataset. Due to the large parallax between different camera views, we choose 2D human keypoints estimated from 20 consecutive frames as image feature points to solve the camera pose parameters. We compare our camera estimation with Bundle Adjustment\cite{triggs2000bundle} and RANSAC 8-point algorithm\cite{sur2008computing}. Bundle Adjustment choose to minimize the reprojection error using the LM algorithm to find the camera parameters and the RANSAC 8-point algorithm uses the 8-point method to find the fundamental matrix after screening out the outliers by RANSAC, and then uses the eigenvalue decomposition to find the camera pose. We provide some visualization examples in Figure \ref{fig:3}. From the Figure \ref{fig:3}, we can see that Bundle Adjustment often falls into the local optimal solution, resulting in poor camera poses. The RANSAC 8-point algorithm can remove outliers, but the estimation of the camera pose is sensitive to 2D human pose estimation, which will lead to accumulated errors when reconstructing 3D human poses. The 3D human and camera pose estimation results are more robust and accurate when introducing our Probabilistic Triangulation (see (c) of Figure \ref{fig:3})

Table \ref{tab:4} shows more detailed results of camera pose evaluation, using the same experimental setup as \cite{bartol2022generalizable}. Among the evaluation metrics used: 1. 3D estimation error(mm) $E_{3D} = E[\| \hat{x^{3D}} - x^{3D} \|_2]$ ; 2. Projection 2D error(mm)$E_{2D} = E[\|\hat{x^{2D}} - x^{2D} \|_2]$; 3. Camera rotation error $E_{R} = E[\| \hat{q} - q \|_2]$; 4. Camera translation error (mm) $E_{t} = E[\| \hat{t} - t \|_2] $.

\begin{figure*}
  \centering
    \includegraphics[width=17cm]{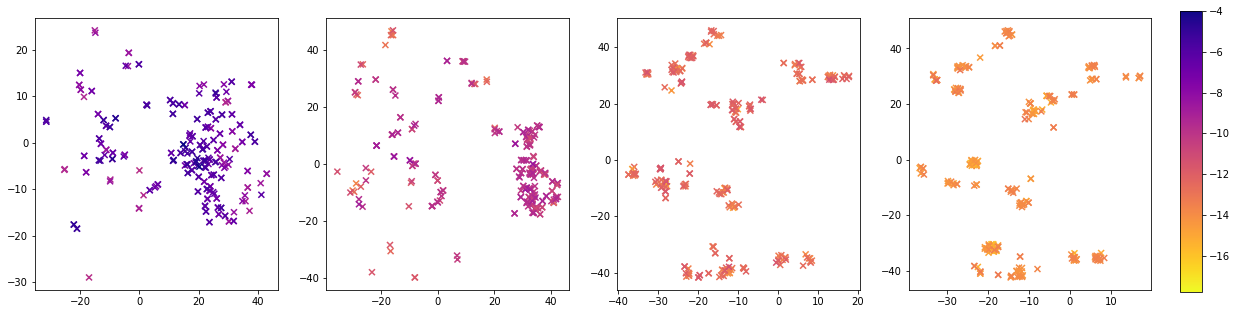}
  \caption{Visualization of the sampled points in four iterations after dimensionality reduction using t-SNE. From left to right corresponds to the sampled values after four iterations. The value of the sampled point is the logarithmic sampling loss, and the smaller the value means that the sampled point is closer to the true value.}
  \label{fig:4}
\end{figure*}

\begin{table}[h]
  \centering
  \tabcolsep=0.08cm
  \begin{tabular}{c|cccc}
    \toprule
    methods & $E_{3D}\downarrow$ &$E_{2D}\downarrow$ & $E_{R}\downarrow$ & $E_{t}\downarrow$\\
    \midrule
    BA  & 736.82 & 130.18 & 0.025 & 46.87 \\
    RANSAC 8-point  & 97.50 & 50.14 & 0.012  & 5.30\\
    Our methods   & \textbf{27.80} & \textbf{5.98} & \textbf{0.0043} &  \textbf{2.71} \\
    \bottomrule
  \end{tabular}
  \caption{Quantitative experiments on camera pose estimation on the Human3.6M dataset.}
  \label{tab:4}
  \vspace{-13pt}
\end{table}

\subsection{Cross-dataset experiments}

\begin{table}
  \centering
  \tabcolsep=0.2cm
  \begin{tabular}{c|cccc}
    \toprule
    Method & Train & Self-Test & H36M-Test & diff\\
    \midrule
    Bartol et al.\cite{bartol2022generalizable} & CMU1 & 25.8 & 33.5 & 30\% \\
    & CMU2 & 26.0 & 33.4 & 28\% \\
    & CMU3 & 25.0 & 31.0 & 24\% \\
    & CMU4 & 25.1 & 32.5 & 28\% \\
    \midrule
    Ours method & CMU1 & 24.6 & 30.8 & 25\% \\
    & CMU2 & 23.7 & 29.1 & 23\% \\
    & CMU3 & 24.1 & 29.2 & 21\% \\
    & CMU4 & 24.9 & 31.1 & 25\% \\ 
    \bottomrule
  \end{tabular}
  \caption{Experiments on generalization ability between datasets (MPJPE in mm). The CMU Panoptic dataset was split into four sub-datasets with different number of cameras and different spatial distributions of cameras. The models were trained on the four datasets separately, and the results were evaluated on their own test set and the Human3.6M test set. The last column represents the cross-data accuracy decline rate.}
  \label{tab:3}
\end{table}

To test the generalization ability of Probabilistic Triangulation across datasets, we segment CMU Panoptic and Human36M into five sub-datasets in the same way as Bartol et al.\cite{bartol2022generalizable}. Cross-dataset experiments are conducted between these sub-datasets two by two. It was found that there was no significant loss of accuracy between the CMU Panoptic sub-datasets even after segmentation (this is because the distribution between the sub-datasets of the CMU Panoptic dataset is very similar). However, the results trained from the CMU dataset have a significant loss of accuracy on Human3.6M. Table \ref{tab:3} shows the generalization ability of Probabilistic Triangulation and Bartol et al.\cite{bartol2022generalizable} from the CMU dataset to the Human3.6M dataset. Probabilistic Triangulation exhibits stronger generalization.

\subsection{Ablation study on the number of cameras}

We show the results of the ablation study in Table \ref{tab:5}. After training on 4-view data, our method generalizes well to 3-view and 2-view because the network has seen rich information during training. There is some performance degradation when the model is trained on 2-view and tested on 3-view and 4-view as shown in Table \ref{tab:5}. 

\begin{table}[h]
  \centering
  \tabcolsep=0.08cm
  \begin{tabular}{c|ccc}
    \toprule
      & 2-view & 3-view & 4-view\\
    \midrule
    Training on 2-view  & 29.5& 42.4 &  39.2  \\
    Training on 4-view  &  32.6 & 29.3 & 27.8\\
    \bottomrule
  \end{tabular}
  \caption{Ablation study on Human3.6M for camera numbers (MPJPE, mm).}
  \label{tab:5}
  \vspace{-13pt}
\end{table}

\subsection{Ablation study on each module}

In order to demonstrate the irreplaceable role that probabilistic triangularization plays in camera pose estimation and 3D human reconstruction in uncalibrated 3D human pose estimation. We design two ablation experiments to remove the influence of the rest of the network on the results and to demonstrate the role of Probabilistic Triangulation better.

First, we use 2D heatmaps ground truth as the input of Probabilistic Triangulation to eliminate the effect of 2D backbone on the results. Also, the upper bound of our method is shown (see Table \ref{tab:6} middle). The table shows that our method achieves comparable results to the calibration method \cite{iskakov2019learnable} when given the 2D ground truth. 

The 3D error still cannot reach zero under the condition of using 2D ground truth because there is a theoretical upper limit to the accuracy of the voxel structure. However, the 2D human pose estimator error cannot be avoided, and the better robustness of the voxel structure to error is the reason it was chosen.

In addition, we demonstrate the necessity of introducing the camera pose estimation module by comparing it with Iskaov et al. (V2V network and triangulation). We rotated the world coordinate system(0° to 180° randomly around the vertical axis and 0° to 45° randomly around the pitch) to change the values of the camera's external parameters, resulting in a new test set called "Test*" in Table \ref{tab:6}. We can find that only using a V2V network and triangulation will make the network overfit the camera poses, resulting in performance degradation. Our method does not require the input of camera poses, so changes in the world coordinate system do not affect our method's final pose estimation results. It demonstrates that our camera estimation module, i.e., the main contribution, is necessary and reasonable.

\begin{table}[h]
  \centering
  \tabcolsep=0.30cm
  \begin{tabular}{c|cc | c}
    \toprule
      & Baseline & 2D GT & Test*\\
    \midrule
    Iskakov et al. & 20.8 & 16.3 & 25.12\\
    Our methods   &  27.8 & 18.5 & 27.8\\
    \bottomrule
  \end{tabular}
  \caption{Pose estimation results (MPJPE, mm) on different 2D and test settings. The middle two columns show the results of 3D pose estimation using the 2D estimator and 2D GT, respectively. The rightmost column shows the estimation results on the Test* set.}
  \label{tab:6}
  \vspace{-13pt}
\end{table}

\subsection{Visualization of Sampling Process}

In order to visualize the sampling process in the Probabilistic Triangulation, the 256 sampling points of each iteration are embedded by t-SNE and plotted in Figure \ref{fig:4}. From the figure, as the iteration proceeds, the approximation of the camera distribution becomes more and more accurate, the sampled points are closer to the ground truth, and the sampling loss keeps decreasing. At the same time, the whole camera pose space is not smooth, and there are many local optimums and pseudo-solutions. The 2D human pose estimator generates noise, making the classical camera pose estimation more likely to fall into local optimal solutions. Distributed representation of camera pose (Probabilistic Triangulation) can alleviate this problem.

Furthermore, from Figure \ref {fig:4}, the sampling points are concentrated around a few extreme points as the iteration parameters increase. This is because an incremental solution exists to the camera pose estimation, so there are multiple extreme points. However, only one of them has real physical significance, whereas, in the projection process, these extreme points produce the same projection result. The distribution also tends to converge to a single extreme point if an initial value is used to initialize the camera pose distribution, and the sampling points of the distribution are spread out to appear near multiple extreme points if a uniform distribution is used to initialize the distribution.

\subsection{Generalize to Wide Scenes}

\begin{figure}
  \centering
    \includegraphics[width=7cm]{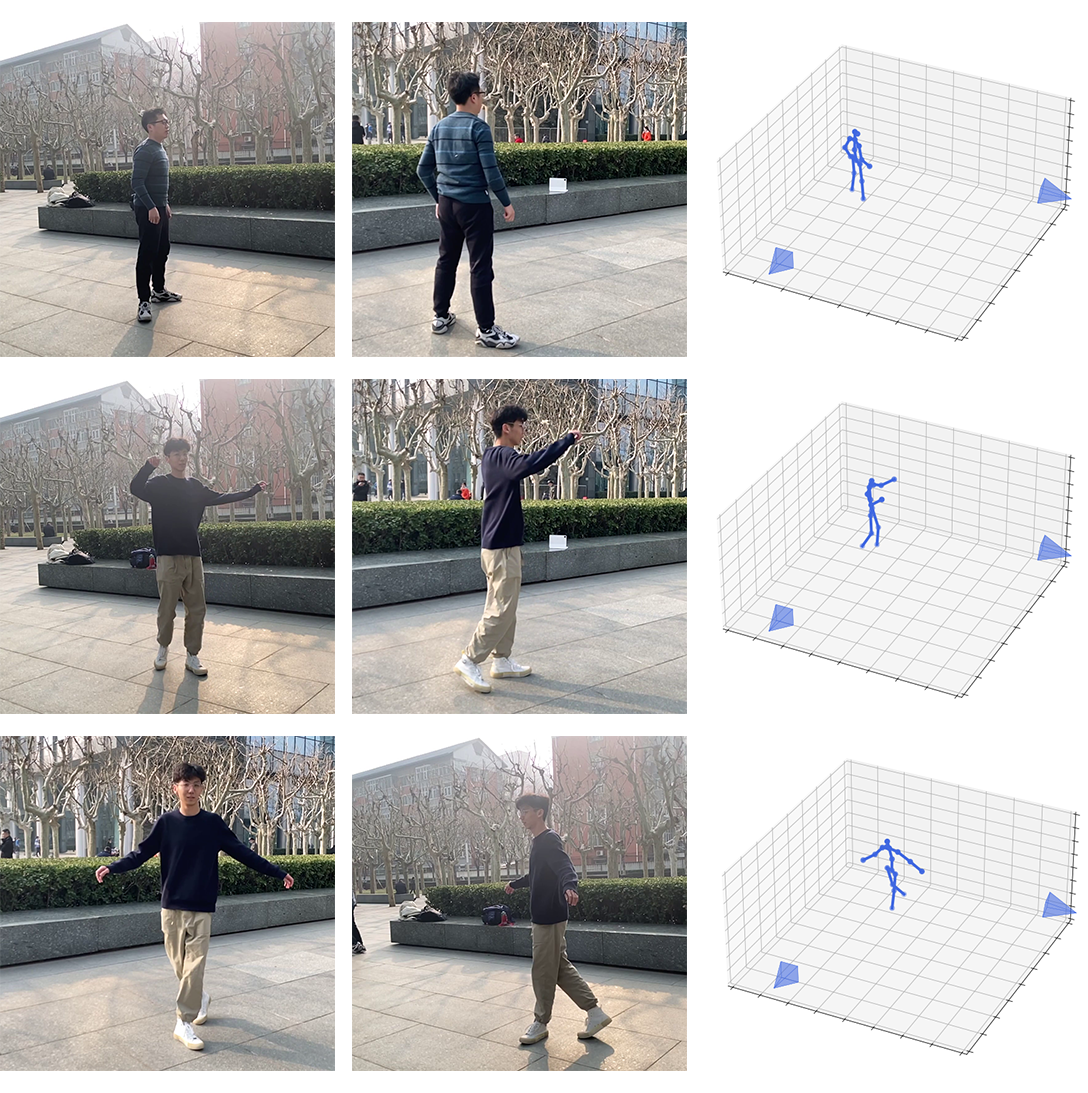}
  \caption{Our method demonstrates the results in a non-calibrated wild scene. The first and second columns represent the two iPhone camera views, and the third column shows the estimated camera pose and 3D human pose.}
  \label{fig:5}
\end{figure}

To demonstrate the generalization of our Probabilistic Triangulation in the wild scenes, we use a model trained from the CMU Panoptic dataset and test it on the natural scenes. Two iPhones with a 26mm primary camera were used for the shooting, and the intrinsic parameters were read using the swift function. The results are shown in Figure \ref{fig:5}. Our method remains in effect even when the camera space location and the number of cameras change.

\section{Conclusions and limitations}
We propose a novel Probabilistic Triangulation module for multi-view 3D human pose estimation, which enables the existing calibrated methods to reduce the dependence on calibration and can be applied to uncalibrated wild scenes. Extensive experiments demonstrate that the 3D human pose estimation based on our  Probabilistic Triangulation can achieve state-of-the-art performance when compared to uncalibrated methods. And the employment of camera pose distribution and Monte Carlo sampling alleviates the problem of easy access to local optimums and pseudo-solutions in camera pose estimation.  Since our Probabilistic Triangulation does not take the camera pose parameters as input, it is possible to build the multi-view 3D human pose estimation system in wild scenes and achieve a high-accurate capture. In the wild scenes, the Probabilistic Triangulation is not affected by the spatial location changes and the number of cameras. It can be directly applied to uncalibrated scenes after training with calibrated data. 

The main limitation of our method is that it can only be applied to single-person scenes. Extending to multi-person scenarios requires taking into account the cross-view matching problem, which is difficult to solve in the absence of a camera pose. In addition, although the Probabilistic Triangulation does not require the camera pose as input during validation, the ground truth of the 3D human pose and the camera pose are used in the training process, and these data need to be obtained from the dataset of the calibration scene. In fact, the ground truth of camera pose is not directly involved in the training, but indirectly guides the 2D pose estimation network learning through cross-view information. The ground truth of the 3D human pose influences the 2D pose estimation network and enables the 3D pose reconstruction network to learn the human prior and integrate the information generated by previous sampling.
\\
\\

\hspace{-0.5cm}\textbf{\large{Acknowledgements}} This work was supported by National Key Research and Development Program of China (NO. 2022YFB3303202).

\newpage

{\small
\bibliographystyle{ieee_fullname}
\bibliography{main}

\begin{thebibliography}{10}\itemsep=-1pt

\bibitem{agarwal2015others}
Sameer Agarwal and Keir Mierle.
\newblock Others,“ceres solver,”.
\newblock {\em Avaliable: http://ceres-solver. org}, 2015.

\bibitem{bartol2022generalizable}
Kristijan Bartol, David Bojani{\'c}, Tomislav Petkovi{\'c}, and Tomislav
  Pribani{\'c}.
\newblock Generalizable human pose triangulation.
\newblock In {\em Proceedings of the IEEE/CVF Conference on Computer Vision and
  Pattern Recognition}, pages 11028--11037, 2022.

\bibitem{belagiannis20143d}
Vasileios Belagiannis, Sikandar Amin, Mykhaylo Andriluka, Bernt Schiele, Nassir
  Navab, and Slobodan Ilic.
\newblock 3d pictorial structures for multiple human pose estimation.
\newblock In {\em Proceedings of the IEEE Conference on Computer Vision and
  Pattern Recognition}, pages 1669--1676, 2014.

\bibitem{borodulina2019application}
Anastasiia Borodulina.
\newblock Application of 3d human pose estimation of motion capture and
  character animation.
\newblock 2019.

\bibitem{bultmann2021real}
Simon Bultmann and Sven Behnke.
\newblock Real-time multi-view 3d human pose estimation using semantic feedback
  to smart edge sensors.
\newblock {\em arXiv preprint arXiv:2106.14729}, 2021.

\bibitem{cao2017realtime}
Zhe Cao, Tomas Simon, Shih-En Wei, and Yaser Sheikh.
\newblock Realtime multi-person 2d pose estimation using part affinity fields.
\newblock In {\em Proceedings of the IEEE conference on computer vision and
  pattern recognition}, pages 7291--7299, 2017.

\bibitem{chen2020end}
Bo Chen, Alvaro Parra, Jiewei Cao, Nan Li, and Tat-Jun Chin.
\newblock End-to-end learnable geometric vision by backpropagating pnp
  optimization.
\newblock In {\em Proceedings of the IEEE/CVF Conference on Computer Vision and
  Pattern Recognition}, pages 8100--8109, 2020.

\bibitem{chen2021monorun}
Hansheng Chen, Yuyao Huang, Wei Tian, Zhong Gao, and Lu Xiong.
\newblock Monorun: Monocular 3d object detection by reconstruction and
  uncertainty propagation.
\newblock In {\em Proceedings of the IEEE/CVF Conference on Computer Vision and
  Pattern Recognition}, pages 10379--10388, 2021.

\bibitem{chen2022epro}
Hansheng Chen, Pichao Wang, Fan Wang, Wei Tian, Lu Xiong, and Hao Li.
\newblock Epro-pnp: Generalized end-to-end probabilistic perspective-n-points
  for monocular object pose estimation.
\newblock In {\em Proceedings of the IEEE/CVF Conference on Computer Vision and
  Pattern Recognition}, pages 2781--2790, 2022.

\bibitem{chen2018cascaded}
Yilun Chen, Zhicheng Wang, Yuxiang Peng, Zhiqiang Zhang, Gang Yu, and Jian Sun.
\newblock Cascaded pyramid network for multi-person pose estimation.
\newblock In {\em Proceedings of the IEEE conference on computer vision and
  pattern recognition}, pages 7103--7112, 2018.

\bibitem{cheng20203d}
Yu Cheng, Bo Yang, Bo Wang, and Robby~T Tan.
\newblock 3d human pose estimation using spatio-temporal networks with explicit
  occlusion training.
\newblock In {\em Proceedings of the AAAI Conference on Artificial
  Intelligence}, volume~34, pages 10631--10638, 2020.

\bibitem{dong2019fast}
Junting Dong, Wen Jiang, Qixing Huang, Hujun Bao, and Xiaowei Zhou.
\newblock Fast and robust multi-person 3d pose estimation from multiple views.
\newblock In {\em Proceedings of the IEEE/CVF Conference on Computer Vision and
  Pattern Recognition}, pages 7792--7801, 2019.

\bibitem{du2017rpan}
Wenbin Du, Yali Wang, and Yu Qiao.
\newblock Rpan: An end-to-end recurrent pose-attention network for action
  recognition in videos.
\newblock In {\em Proceedings of the IEEE international conference on computer
  vision}, pages 3725--3734, 2017.

\bibitem{gordon2021flex}
Brian Gordon, Sigal Raab, Guy Azov, Raja Giryes, and Daniel Cohen-Or.
\newblock Flex: Parameter-free multi-view 3d human motion reconstruction.
\newblock {\em arXiv preprint arXiv:2105.01937}, 2021.

\bibitem{hartley2003multiple}
Richard Hartley and Andrew Zisserman.
\newblock {\em Multiple view geometry in computer vision}.
\newblock Cambridge university press, 2003.

\bibitem{he2020epipolar}
Yihui He, Rui Yan, Katerina Fragkiadaki, and Shoou-I Yu.
\newblock Epipolar transformer for multi-view human pose estimation.
\newblock In {\em Proceedings of the IEEE/CVF Conference on Computer Vision and
  Pattern Recognition Workshops}, pages 1036--1037, 2020.

\bibitem{ionescu2013human3}
Catalin Ionescu, Dragos Papava, Vlad Olaru, and Cristian Sminchisescu.
\newblock Human3. 6m: Large scale datasets and predictive methods for 3d human
  sensing in natural environments.
\newblock {\em IEEE transactions on pattern analysis and machine intelligence},
  36(7):1325--1339, 2013.

\bibitem{iskakov2019learnable}
Karim Iskakov, Egor Burkov, Victor Lempitsky, and Yury Malkov.
\newblock Learnable triangulation of human pose.
\newblock In {\em Proceedings of the IEEE/CVF International Conference on
  Computer Vision}, pages 7718--7727, 2019.

\bibitem{joo2015panoptic}
Hanbyul Joo, Hao Liu, Lei Tan, Lin Gui, Bart Nabbe, Iain Matthews, Takeo
  Kanade, Shohei Nobuhara, and Yaser Sheikh.
\newblock Panoptic studio: A massively multiview system for social motion
  capture.
\newblock In {\em Proceedings of the IEEE International Conference on Computer
  Vision}, pages 3334--3342, 2015.

\bibitem{lee2022extrinsic}
Sang-Eun Lee, Keisuke Shibata, Soma Nonaka, Shohei Nobuhara, and Ko Nishino.
\newblock Extrinsic camera calibration from a moving person.
\newblock {\em IEEE Robotics and Automation Letters}, 7(4):10344--10351, 2022.

\bibitem{liu2022recent}
Wu Liu and Tao Mei.
\newblock Recent advances of monocular 2d and 3d human pose estimation: A deep
  learning perspective.
\newblock {\em ACM Computing Surveys (CSUR)}, 2022.

\bibitem{ma2021transfusion}
Haoyu Ma, Liangjian Chen, Deying Kong, Zhe Wang, Xingwei Liu, Hao Tang, Xiangyi
  Yan, Yusheng Xie, Shih-Yao Lin, and Xiaohui Xie.
\newblock Transfusion: Cross-view fusion with transformer for 3d human pose
  estimation.
\newblock {\em arXiv preprint arXiv:2110.09554}, 2021.

\bibitem{munea2020progress}
Tewodros~Legesse Munea, Yalew~Zelalem Jembre, Halefom~Tekle Weldegebriel,
  Longbiao Chen, Chenxi Huang, and Chenhui Yang.
\newblock The progress of human pose estimation: a survey and taxonomy of
  models applied in 2d human pose estimation.
\newblock {\em IEEE Access}, 8:133330--133348, 2020.

\bibitem{obdrvzalek2012real}
{\v{S}}t{\v{e}}p{\'a}n Obdr{\v{z}}{\'a}lek, Gregorij Kurillo, Jay Han, Ted
  Abresch, and Ruzena Bajcsy.
\newblock Real-time human pose detection and tracking for tele-rehabilitation
  in virtual reality.
\newblock In {\em Medicine Meets Virtual Reality 19}, pages 320--324. IOS
  Press, 2012.

\bibitem{pavllo20193d}
Dario Pavllo, Christoph Feichtenhofer, David Grangier, and Michael Auli.
\newblock 3d human pose estimation in video with temporal convolutions and
  semi-supervised training.
\newblock In {\em Proceedings of the IEEE/CVF Conference on Computer Vision and
  Pattern Recognition}, pages 7753--7762, 2019.

\bibitem{qiu2019cross}
Haibo Qiu, Chunyu Wang, Jingdong Wang, Naiyan Wang, and Wenjun Zeng.
\newblock Cross view fusion for 3d human pose estimation.
\newblock In {\em Proceedings of the IEEE/CVF International Conference on
  Computer Vision}, pages 4342--4351, 2019.

\bibitem{regazzoni2014rgb}
Daniele Regazzoni, Giordano De~Vecchi, and Caterina Rizzi.
\newblock Rgb cams vs rgb-d sensors: Low cost motion capture technologies
  performances and limitations.
\newblock {\em Journal of Manufacturing Systems}, 33(4):719--728, 2014.

\bibitem{remelli2020lightweight}
Edoardo Remelli, Shangchen Han, Sina Honari, Pascal Fua, and Robert Wang.
\newblock Lightweight multi-view 3d pose estimation through camera-disentangled
  representation.
\newblock In {\em Proceedings of the IEEE/CVF conference on computer vision and
  pattern recognition}, pages 6040--6049, 2020.

\bibitem{simon2017hand}
Tomas Simon, Hanbyul Joo, Iain Matthews, and Yaser Sheikh.
\newblock Hand keypoint detection in single images using multiview
  bootstrapping.
\newblock In {\em Proceedings of the IEEE conference on Computer Vision and
  Pattern Recognition}, pages 1145--1153, 2017.

\bibitem{sur2008computing}
Fr{\'e}d{\'e}ric Sur, Nicolas Noury, and Marie-Odile Berger.
\newblock Computing the uncertainty of the 8 point algorithm for fundamental
  matrix estimation.
\newblock In {\em 19th British Machine Vision Conference-BMVC 2008}, page~10,
  2008.

\bibitem{takahashi2018human}
Kosuke Takahashi, Dan Mikami, Mariko Isogawa, and Hideaki Kimata.
\newblock Human pose as calibration pattern; 3d human pose estimation with
  multiple unsynchronized and uncalibrated cameras.
\newblock In {\em Proceedings of the IEEE Conference on Computer Vision and
  Pattern Recognition Workshops}, pages 1775--1782, 2018.

\bibitem{triggs2000bundle}
Bill Triggs, Philip~F McLauchlan, Richard~I Hartley, and Andrew~W Fitzgibbon.
\newblock Bundle adjustment—a modern synthesis.
\newblock In {\em Vision Algorithms: Theory and Practice: International
  Workshop on Vision Algorithms Corfu, Greece, September 21--22, 1999
  Proceedings}, pages 298--372. Springer, 2000.

\bibitem{tu2020voxelpose}
Hanyue Tu, Chunyu Wang, and Wenjun Zeng.
\newblock Voxelpose: Towards multi-camera 3d human pose estimation in wild
  environment.
\newblock In {\em European Conference on Computer Vision}, pages 197--212.
  Springer, 2020.

\bibitem{tyler1987statistical}
David~E Tyler.
\newblock Statistical analysis for the angular central gaussian distribution on
  the sphere.
\newblock {\em Biometrika}, 74(3):579--589, 1987.

\bibitem{xiang2019monocular}
Donglai Xiang, Hanbyul Joo, and Yaser Sheikh.
\newblock Monocular total capture: Posing face, body, and hands in the wild.
\newblock In {\em Proceedings of the IEEE/CVF conference on computer vision and
  pattern recognition}, pages 10965--10974, 2019.

\bibitem{zhang2021direct}
Jianfeng Zhang, Yujun Cai, Shuicheng Yan, Jiashi Feng, et~al.
\newblock Direct multi-view multi-person 3d pose estimation.
\newblock {\em Advances in Neural Information Processing Systems},
  34:13153--13164, 2021.

\bibitem{zhang2022mixste}
Jinlu Zhang, Zhigang Tu, Jianyu Yang, Yujin Chen, and Junsong Yuan.
\newblock Mixste: Seq2seq mixed spatio-temporal encoder for 3d human pose
  estimation in video.
\newblock In {\em Proceedings of the IEEE/CVF Conference on Computer Vision and
  Pattern Recognition}, pages 13232--13242, 2022.

\bibitem{zheng20213d}
Ce Zheng, Sijie Zhu, Matias Mendieta, Taojiannan Yang, Chen Chen, and Zhengming
  Ding.
\newblock 3d human pose estimation with spatial and temporal transformers.
\newblock In {\em Proceedings of the IEEE/CVF International Conference on
  Computer Vision}, pages 11656--11665, 2021.

\bibitem{zhu2022motionbert}
Wentao Zhu, Xiaoxuan Ma, Zhaoyang Liu, Libin Liu, Wayne Wu, and Yizhou Wang.
\newblock Motionbert: Unified pretraining for human motion analysis.
\newblock {\em arXiv preprint arXiv:2210.06551}, 2022.

\end{thebibliography}
}

\end{document}